\documentclass[10pt,twocolumn,letterpaper]{article}

\usepackage{cvpr}
\usepackage{times}
\usepackage{epsfig}
\usepackage{graphicx}
\usepackage{amsmath}
\usepackage{amssymb}
\usepackage{array}
\usepackage{dsfont}
\usepackage{multirow}
\usepackage{subfigure}
\usepackage{multirow}
\usepackage{color}
\definecolor{gray}{rgb}{0.5,0.5,0.5}
\iftrue
	\newcommand{\topic}[1]{\textcolor{gray}{\textbf{(#1.)}}}
	\newcommand{\outline}[1]{{\textcolor{blue}{[[{#1}]]}}}
	\newcommand{\commenttext}[1]{\textcolor{red}{[[{#1}]]}}
	\newcommand{\commentfoot}[1]{\footnote{\textcolor{red}{\textit{#1}}}}
\else 
	\newcommand{\topic}[1]{}
	\newcommand{\outline}[1]{}
	\newcommand{\commenttext}[1]{}
	\newcommand{\commentfoot}[1]{}
\fi

\newcommand{\cutcaptiondown}{\vspace*{-0.12in}}
\usepackage[pagebackref=true,breaklinks=true,letterpaper=true,colorlinks,bookmarks=false]{hyperref}
\usepackage{algpseudocode}
\usepackage{algorithm}
\graphicspath{{./images/}}

\newcommand{\myparagraph}[1]{\vspace{2pt}\noindent{\bf #1}}
\newcommand{\SJElong}{Structured Joint Embedding\xspace}
\newcommand{\SJE}{SJE\xspace}

\def\D{\Delta}
\def\Re{\mathbb R}
\def\p{\varphi}
\def\1{\mathds{1}}

\makeatletter
\DeclareRobustCommand\onedot{\futurelet\@let@token\@onedot}
\def\@onedot{\ifx\@let@token.\else.\null\fi\xspace}

\def\eg{{e.g}\onedot} 
\def\ie{{i.e}\onedot} 
 
\def\etc{{etc}\onedot} 
\def\wrt{w.r.t\onedot}

\g@addto@macro\normalsize{%
  \setlength\abovedisplayskip{7pt}
  \setlength\belowdisplayskip{7pt}
  \setlength\abovedisplayshortskip{5pt}
  \setlength\belowdisplayshortskip{5pt}
}

\makeatother

\usepackage{framed}
\usepackage{xcolor}
\definecolor{gainsboro}{RGB}{220,220,220}

\newcommand{\X}{\mathcal{X}}
\newcommand{\Y}{\mathcal{Y}}

\cvprfinalcopy 

\def\cvprPaperID{1690} 

\ifcvprfinal\fi
\begin{document}

\title{Evaluation of Output Embeddings for Fine-Grained Image Classification}

\author{Zeynep Akata$^*$, Scott Reed$^\dagger$, Daniel Walter$^\dagger$, Honglak Lee$^\dagger$ and Bernt Schiele$^*$ \vspace{4mm} \\ 
{\small
\begin{tabular}{cp{1cm}c}
$^*$ Computer Vision and Multimodal Computing & & $^\dagger$ Computer Science and Engineering Division \\
 Max Planck Institute for Informatics, Saarbrucken, Germany & & University of Michigan, Ann Arbor \\
\end{tabular}
}
}

\maketitle

\begin{abstract}
Image classification has advanced significantly in recent years with the availability of large-scale image sets.
However, fine-grained classification remains a major challenge due to the annotation cost of large numbers of fine-grained categories.
This project shows that compelling classification performance can be achieved on such categories even without labeled training data. 
Given image and class embeddings, we learn a compatibility function such that matching embeddings are assigned a higher score than mismatching ones; zero-shot classification of an image proceeds by finding the label yielding the highest joint compatibility score.
We use state-of-the-art image features and focus on different supervised attributes and unsupervised output embeddings either derived from hierarchies or learned from unlabeled text corpora.
We establish a substantially improved state-of-the-art on the Animals with Attributes and Caltech-UCSD Birds datasets.
Most encouragingly, we demonstrate that purely unsupervised output embeddings (learned from Wikipedia and improved with fine-grained text) achieve compelling results, even outperforming the previous supervised state-of-the-art.
By combining different output embeddings, we further improve results.
\end{abstract}


\vspace*{-0.15in}

\section{Introduction}

The image classification problem has been redefined by the emergence of large scale datasets such as ImageNet~\cite{DDS09}. 
Since deep learning methods~\cite{KSH12} dominated recent Large-Scale Visual Recognition Challenges (ILSVRC12-14), the attention of the computer vision community has been drawn to Convolutional Neural Networks (CNN)~\cite{LBBH98}.
Training CNNs requires massive amounts of labeled data; but, in fine-grained image collections, where the categories are visually very similar, the data population decreases significantly.
We are interested in the most extreme case of learning with a limited amount of labeled data, zero-shot learning, in which no labeled data is available for some classes.

Without labels, we need alternative sources of information that relate object classes. 
Attributes~\cite{FZ07, FEHF09, LNH13}, which describe well-known common characteristics of objects, are an appealing source of information, and they can be easily obtained through crowd-sourcing techniques~\cite{DKF13, PG11}. %
However, fine-grained concepts present a special challenge: due to the high degree of similarity among categories, a large number of attributes are required to effectively model these subtle differences.
This increases the cost of attribute annotation. 
One aim of this work is to move towards 
eliminating the human labeling component from zero-shot learning, \eg by using alternative sources of information. 
\begin{figure}[t]
\begin{center}
\includegraphics[width=\linewidth, trim=30 40 30 40]{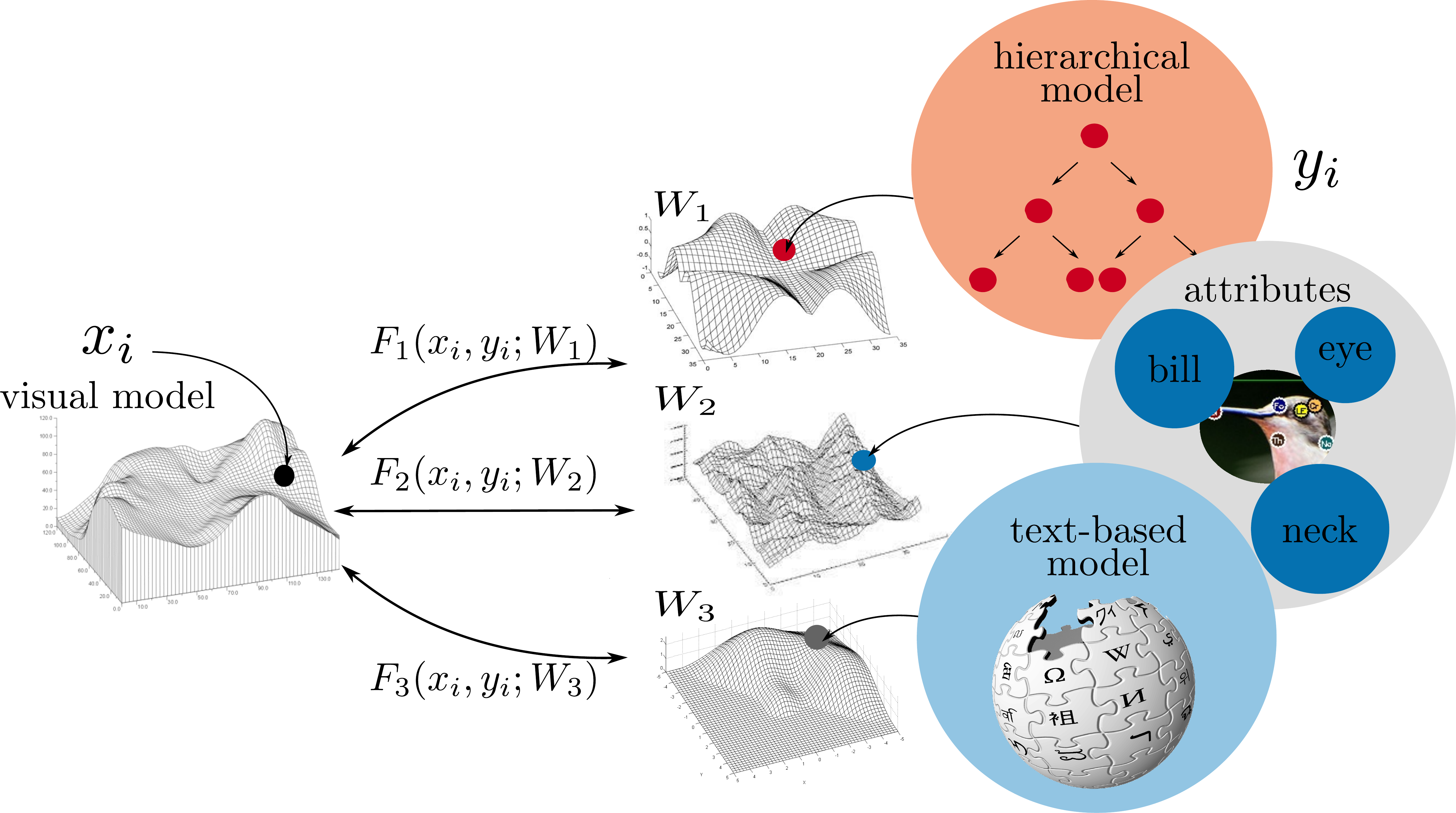}
\end{center}
\caption{\SJElong leverages images ($x_i$) and labels ($y_i$) by learning parameters $W$ of a function $F(x_i, y_i, W)$ that measures the compatibility between input ($\theta(x_i)$) and output embeddings ($\p(y_i)$). It is a general framework that can be applied to any learning problem with more than one modality.} \vspace{-2mm}
\label{fig:methodology}
\cutcaptiondown
\end{figure}

On the other hand, large-margin support vector machines (SVM) operate with labeled training images, so a lack of labels limits their use for this task.
Inspired by previous work on label embedding~\cite{WBU10, BWG10, APHS13} and structured SVMs~\cite{TJH05,NL11}, we propose to use a \SJElong (\SJE) framework (Fig.~\ref{fig:methodology}) that relates input embeddings (\ie image features) and output embeddings (\ie side information) through a compatibility function, therefore taking advantage of a structure in the output space. The \SJE framework separates the subspace learning problem from the specific input and output features used in a given application. As a general framework, it can be applied to any learning problem where more than one modality is provided for an object. 

Our contributions are: (1) We demonstrate that unsupervised class embeddings trained from large unlabeled text corpora are competitive to previously published results that use human supervision. (2) Using the most recent deep architectures as input embeddings, we significantly improve the state-of-the-art (SoA). (3) We extensively evaluate several unsupervised output embeddings for fine-grained classification in a zero-shot setting on three challenging datasets. (4) 
By combining different output embeddings we obtain best results, surpassing the SoA by a large margin. 
(5) We propose a novel weakly-supervised Word2Vec variant that improves the accuracy when combined with other output embeddings.  

The rest of the paper is organized as follows. Section~\ref{sec:prev} provides a review of the relevant literature; Sec.~\ref{sec:model} details the \SJE method; Sec.~\ref{sec:output_emb} explains the output embeddings that we analyze; Sec.~\ref{sec:exp} presents our experimental evaluation; Sec.~\ref{sec:conclusion} presents the discussion and our conclusions.


\section{Related Work }
\label{sec:prev}
Learning to classify in the absence of labeled data (zero-shot learning)~\cite{YA10,RSS11,KKTH12,LNH13,APHS15,NMBSSFCD13,MGS14, FHXFG14} is a challenging problem, and achieving better-than-chance performance requires structure in the output space.
Attributes~\cite{FZ07, FEH10, LNH13} provide one such space; they relate different classes through well-known and shared characteristics of objects. 
Attributes, which are often collected manually~\cite{KKTH12, PG11, DPCG12}, have shown promising results in various applications, \ie caption generation~\cite{KPD11,OKB11}, face recognition~\cite{SKBB12,CGG13}, image retrieval~\cite{SFD11,DRS11}, action recognition~\cite{LKS11,YJKLGL11} and image classification~\cite{LNH13, APHS15}.
The main challenge of attribute-based zero-shot learning arises on more challenging fine-grained data collections~\cite{CaltechUCSDBirdsDataset, StanfordDogsDataset}, in which categories may visually differ only subtly.
Therefore, generic attributes fail at modeling small intra-class variance between objects. Improved performance requires a large number of specific attributes which increases the cost of data gathering.

As an alternative to manual annotation, side information can be collected automatically from text corpora. Bag-of-words~\cite{H54} is an example where class embeddings correspond to histograms of vocabulary words extracted automatically from unlabeled text. Another example is using taxonomical order of classes~\cite{TJH05} as structured output embeddings. Such a taxonomy can be built automatically from a pre-defined ontology such as WordNet~\cite{WordNet,RSS10,APHS13} . In this case, the distance between nodes is measured using semantic similarity metrics~\cite{JC97, LC94, Lin98, Res95}. Finally, distributed text representations~\cite{MSCCD13, PSM14} learned from large unsupervised text corpora can be employed as structured embeddings. We compare several representatives of these methods (and their combinations) in our evaluation.

Embedding labels in an Euclidean space is an effective tool to model latent relationships between classes~\cite{BWG10}. These relationships can be collected separately from the data~\cite{HKL09, DB95}, learned from the data~\cite{WBU10,Hastie:Tibshirani:Friedman:2008} or derived from side information~\cite{FZ07, FCSBM13, APHS13, NMBSSFCD13}. In order to collect relationships independently of data, compressed sensing~\cite{HKL09} uses random projections whereas Error Correcting Output Codes~\cite{DB95} builds embeddings inspired from information theory. WSABIE~\cite{WBU10} uses images with their corresponding labels to learn an embedding of the labels, and CCA~\cite{Hastie:Tibshirani:Friedman:2008} maximizes the correlation between two different data modalities. DeViSE~\cite{FCSBM13} employs a ranking formulation for zero-shot learning using images and distributed text representations. The ALE~\cite{APHS13} method employs an approximate ranking formulation for the same using images and attributes. ConSe~\cite{NMBSSFCD13}  uses the probabilities of a softmax-output layer to weigh the semantic vectors of all the classes. 
In this work, we use the multiclass objective to learn structured output embeddings obtained from various sources.

Among the closest related work, ALE~\cite{APHS13} uses Fisher Vectors (FV~\cite{PD07}) as input and binary attributes / hierarchies as output embeddings. Similarly, DeviSe~\cite{FCSBM13} uses CNN~\cite{KSH12} features as input and Word2Vec~\cite{MSCCD13} representations as output embeddings. In this work, we benefit from both ideas: (1) We use SoA image features, \ie FV and CNN, (2) among others, we also use attributes and Word2Vec as output embeddings. Our work differs from \cite{FCSBM13} \wrt two aspects: (1) We propose and evaluate several output embedding methods specifically built for fine-grained classification. (2) We show how some of these output embeddings complement each other for zero-shot learning on general and fine-grained datasets. The reader should be aware of~\cite{APHS15}.


\section{{\SJElong}s}
\label{sec:model}

In this work, we aim to leverage input and output embeddings in a joint framework by learning a compatibility between these embeddings. We are interested in the problem of zero-shot learning for image classification where training and test images belong to two disjoint sets of classes. 

Following~\cite{APHS13}, given input/output $x_n \in \X$ and $y_n \in \Y$ from ${\cal S}=\{(x_n,y_n), n =1 \ldots N\}$, \SJElong (\SJE) learns $f: \X \rightarrow \Y$ by minimizing the empirical risk
 $\frac{1}{N} \sum_{n=1}^N \D(y_n,f(x_n))$
where $\D: \Y \times \Y \rightarrow \Re$ defines the cost of predicting $f(x)$ when the true label is $y$. Here, we use the $0/1$ loss.

\subsection{Model} 
\label{subsec:model}
We define a compatibility function $F: \X \times \Y \rightarrow \Re$ between an input space $\X$ and a structured output space $\Y$. Given a specific input embedding, we derive a prediction by maximizing the compatibility $F$ over \SJE as follows:
\begin{equation*}
f(x;w) = \arg \max_{y \in \Y} F(x,y; w).
\label{eq:annot}
\end{equation*}
The parameter vector $w$ can be written as a $D \times E$ matrix $W$ with $D$ being the input embedding dimension and $E$ being the output embedding dimension. This leads to the bi-linear form of the compatibility function:
\begin{equation}
F(x,y;W) = \theta(x)^{\top} W \p(y) . 
\label{eqn:form}
\end{equation}
Here, the input embedding is denoted by $\theta(x)$ and the output embedding by $\p(y)$. The matrix $W$ is learned by enforcing the correct label to be ranked higher than any of the other labels (Sec.~\ref{subsec:learning}), \ie multiclass objective. This formulation is closely related to~\cite{APHS13, FCSBM13, WBU10}. Within the label embedding framework, ALE~\cite{APHS13} and DeViSe~\cite{FCSBM13} use pairwise ranking objective, WSABIE~\cite{WBU10} learns both $\p(y)$ and $W$ through ranking, whereas we use multiclass objective. Similarly, \cite{PPH09,SGSBMN13} use the regression objective and CCA~\cite{Hastie:Tibshirani:Friedman:2008} maximizes the correlation of input and output embeddings. 

\subsection{Parameter Learning} 
\label{subsec:learning}
According to the unregularized structured SVM formulation~\cite{TJH05}, the objective is:
\begin{equation}
\frac{1}{N} \sum_{n=1}^N \max_{y \in \Y}\{0,\ell(x_n,y_n,y)\}.
\label{eqn:ssvm_zsh}
\end{equation}
where the misclassification loss $ \ell(x_n,y_n,y)$ takes the form:
\begin{equation}
\D(y_n,y) + \theta(x_n)^{\top}W\p(y) - \theta(x_n)^{\top}W\p(y_n)
\label{eqn:ssvm_zsh}
\end{equation}
For the zero-shot learning scenario, the training and test classes are disjoint. Therefore, we fix $\p$ to the output embeddings of training classes and learn $W$. For prediction, we project a test image onto the $W$ and search for the nearest output embedding vector (using the dot product similarity) that corresponds to one of the test classes.

We use Stochastic Gradient Descent (SGD) for optimization which consists in sampling $(x_{n},y_{n})$ at each step and searching for the highest ranked class $y$. If $\arg \max_{y \in \Y} \ell(x_n,y_n,y) \neq y_n$, we update $W$ as follows: 
\begin{equation}
 W^{(t)} = W^{(t-1)} + \eta_t \theta(x_n) [\p(y_n) - \p(y)]^{\top} 
\end{equation}
where $\eta_t$ is the learning step-size used at iteration $t$. We use a constant step size chosen by cross-validation and we perform regularization through early stopping.

\subsection{Learning Combined Embeddings}
\label{subsec:comb_emb}
For some classification tasks, there may be multiple output embeddings available, each capturing a different aspect of the structure of the output space. Each may also have a different signal-to-noise ratio. Since each output embedding possibly offers non-redundant information about the output space, as also shown in~\cite{RSS11,APHS15}, we can learn a better joint embedding by combining them together. We model the resulting compatibility score as
\begin{align}
\label{eqn:form_ensemble}
F(x,y;\{W\}_{1..K}) =& \sum_{k} \alpha_{k} \theta(x)^{\top} W_{k} \p_k(y)  \\
\text{s.t.} & \sum_{k}\alpha_{k} = 1\nonumber
\end{align}
where $W_{1},...,W_{K}$ are the joint embedding weight matrices corresponding to the $K$ output embeddings ($\p_k$). In our experiments, we first train each $W_{k}$ independently, then perform a grid search over $\alpha_{k}$ on a validation set. Interestingly, we found that the optimal $\alpha_{k}$ for previously-seen classes is often different from the one for unseen classes. Therefore, it is critical to cross-validate $\alpha_{k}$ on the zero-shot setting.

Note that if we take $\alpha_{k}=1/K, \forall k$, Equation~\ref{eqn:form_ensemble} is equivalent to simply concatenating the $\p_k$. This corresponds to stacking the $W_{k}$ into a single matrix $W$ and computing the standard compatibility as in Equation~\ref{eqn:form}. However, such a stacking learns a large $W$ where a high dimensional $\p$ biases the final prediction. In contrast, $\alpha$ eliminates the bias, leading to better predictions. 
Thus, $\alpha_{k}$ can be thought of as the confidence associated with $\p_k$ whose contribution we can control. We show in Sec.~\ref{subsec:results} that finding an appropriate $\alpha_{k}$ can yield improved accuracy compared to any single $\p$.

\section{Output Embeddings}
\label{sec:output_emb}

In this section, we describe three types of output embeddings: human-annotated attributes, unsupervised word embeddings learned from large text corpora, and hierarchical embeddings derived from WordNet.

\subsection{Embedding by Human Annotation: Attributes}
\label{subsec:attributes}
Annotating images with class labels is a laborious process when the objects represent fine-grained concepts that are not common in our daily lives.
Attributes provide a means to describe such fine-grained concepts.
They model shared characteristics of objects such as color and texture which are easily annotated by humans and converted to machine-readable vector format.
The set of descriptive attributes may be determined by language experts~\cite{LNH13} or by fine-grained object experts~\cite{CaltechUCSDBirdsDataset}.
The association between an attribute and a category can be a binary value depicting the presence/absence of an attribute ($\p^{0,1}$~\cite{LNH13, APHS13, RSS11}) or a continuous value that defines the confidence level of an attribute ($\p^{\cal A}$~\cite{LNH13, APHS15, RSSS12}) for each class.  
We write per-class attributes as:
\begin{equation*}
\p(y) = [\rho_{y,1}, \ldots, \rho_{y,E}]^{\top}
\end{equation*}
where $\rho_{y,i}$ can be $\{0,1\}$ or a real number that associates a class with an attribute, $y$ denotes the associated class and $E$ is the number of attributes. Potentially, $\p^{\cal A}$ encodes more information than $\p^{0,1}$. For instance, for classes \emph{rat, monkey, whale} and the attribute \emph{big}, $\p^{0,1} = [0,0,1]$  implies that in terms of size \emph{rat} $=$ \emph{monkey} $<$ \emph{whale}, whereas $\p^{\cal A} = [2,10,90]$ can be interpreted as \emph{rat} $<$ \emph{monkey} $<<$ \emph{whale} which is more accurate. We empirically show the benefit of $\p^{\cal A}$ over $\p^{0,1}$ in Sec.~\ref{subsec:results}.
In practice, our output embeddings use a per-class vector form, but they can vary in dimensionality ($E$). For the rest of the section we denote the output embeddings as $\p$ for brevity.
%

\subsection{Learning Label Embeddings from Text}
\label{subsec:text}
In this section, we describe unsupervised and weakly-supervised label embeddings mined from text. With these label embeddings, we can (1) avoid dependence on costly manual annotation of attributes and (2) combine the embeddings with attributes, where available, to achieve better performance.

\myparagraph{Word2Vec ($\p^{\cal W}$).} In Word2Vec~\cite{MSCCD13}, a two-layer neural network is trained to predict a set of target words from a set of context words. Words in the vocabulary are assigned with one-shot encoding so that the first layer  acts as a look-up table to retrieve the embedding for any word in the vocabulary. The second layer predicts the target word(s) via hierarchical soft-max. Word2Vec has two main formulations for the target prediction: skip-gram (SG) and continuous bag-of-words (CBOW). In SG, words within a local context window are predicted from the centering word. In CBOW, the center word of a context window is predicted from the surrounding words. Embeddings are obtained by back-propagating the prediction error gradient over a training set of context windows sampled from the text corpus.

\myparagraph{GloVe ($\p^{\cal G}$).} GloVe~\cite{PSM14} incorporates co-occurrence statistics of words that frequently appear together within the document. Intuitively, the co-occurrence statistics encode meaning since semantically similar words such as ``ice'' and ``water'' occur together more frequently than semantically dissimilar words such as ``ice'' and ``fashion.'' The training objective is to learn word vectors such that their dot product equals the co-occurrence probability of these two words. This approach has recently been shown to outperform Word2Vec on the word analogy prediction task~\cite{PSM14}.

\myparagraph{Weakly-supervised Word2Vec ($\p^{{\cal W}_{ws}}$).} The standard Word2Vec~\cite{MSCCD13} scans the entire document using each word within a sample window as the target for prediction. However, if we know the global context, \ie the topic of the document, we can use that topic as our target. For instance, in Wikipedia, the entire article is related to the same topic. Therefore, we can sample our context windows from any location within the article rather than searching for context windows where the topic explicitly appears in the text. We consider this method as a weak form of supervision.

We achieve the best results in our experiments using our novel variant of the CBOW formulation. Here, we pre-train the first layer weights using standard Word2Vec on Wikipedia, and fine-tune the second layer weights using a negative-sampling objective~\cite{goldberg2014word2vec} only on the fine-grained text corpus. These weights correspond to the final output embedding. The negative sampling objective is formulated as follows:
\begin{align}
L &= \sum_{w,c \in D_{+}}\log \sigma (v_{c}^{T}v_{w}) + \sum_{w',c \in D_{-}}\log \sigma (-v_{c}^{T}v_{w'})\\
v_{c} &= \sum_{i \in \text{context(w)}}v_{i} / | \text{context(w)} | \nonumber
\end{align}
where $v_{w}$ and $v_{w'}$ are the label embeddings we seek to learn, and $v_{c}$ is the average of word embeddings $v_{i}$ within a context window around word $w$. 
$D_{+}$ consists of context $v_{c}$ 
and matching targets $v_{w}$, and $D_{-}$ consists of the same $v_{c}$ and mismatching $v_{w'}$. 
To find the $v_{i}$ (which are the columns of the first-layer network weights), we take them from a standard unsupervised Word2Vec model trained on Wikipedia.

During SGD, the $v_{i}$ are fixed and we update each sampled $v_{w}$ and $v_{w'}$ at each iteration.
Intuitively, we seek to maximize the similarity between context and target vectors for matching pairs, and minimize it for mismatching pairs.

\myparagraph{Bag-of-Words ($\p^{\cal B}$).} BoW~\cite{H54} builds a ``bag'' of word frequencies by counting the occurrence of each vocabulary word that appears within a document. It does not preserve the order in which words appear in a document, so it disregards the grammar. We collect Wikipedia articles that correspond to each object class and build a vocabulary of most frequently occurring words. We then build histograms of these words to vectorize our classes.

\subsection{Hierarchical Embeddings}
\label{subsec:hie}
Semantic similarity measures how closely related two word senses are according to their meaning. Such a similarity can be estimated by measuring the distance between terms in an ontology. WordNet\footnote{\texttt{http://wordnetweb.princeton.edu/}}, a large-scale hierarchical database of over 100,000 words for English, provides us a means of building our class hierarchy. 
To measure similarity, we use Jiang-Conrath~\cite{JC97} ($\p^{jcn}$), Lin~\cite{Lin98} ($\p^{lin}$) and path ($\p^{path}$) similarities formulated in Table~\ref{tab:similairty}. We denote our whole family of hierarchical embeddings as $\p^{\cal H}$. For a more detailed survey, the reader may refer to \cite{BHBKP05}.

\begin{table}[t]
 \begin{center}
   \small
{\renewcommand{\arraystretch}{1.5}
  \begin{tabular}{l}
	\hline
	\hline
	 $\rho_{jcn} = 2* IC(mscs(u,v)) - (IC(u) + IC(v))$ \vspace{1mm}\\
	 $\rho_{lin} = \displaystyle{ 2 * IC(mscs(u,v))  \over  {IC(u) + IC(v)} }$  \\
	$\rho_{path}= \displaystyle{ \min_{p\in pth(u,v)} } len(p) $ \vspace{1mm} \\ 
	\hline
	\hline	
  \end{tabular}
} 
 \end{center}
\caption{Notations~\cite{BHBKP05}: mscs (most specific common subsumer), pth (set of paths between two nodes), len (path length), IC (Information Content, defined as the log of the probability of finding a word in a text corpus independent of the hierarchy).} 
\cutcaptiondown
\label{tab:similairty}
\end{table}


\section{Experiments}
\label{sec:exp}

While our main contribution is a detailed analysis of output embeddings, good image representations are crucial to obtain good classification performance. 
In Sec.~\ref{subsec:setting} we detail datasets, input and output embeddings used in our experiments and in Sec.~\ref{subsec:results} we present our results.

\subsection{Experimental Setting}
\label{subsec:setting}

We evaluate \SJE on three datasets: Caltech UCSD Birds (CUB)~\cite{WBPB11} and Stanford Dogs (Dogs)\footnote{We use 113 classes that appear in the Federation Cynologique Internationale (FCI) database of dog breeds.}~\cite{StanfordDogsDataset} are fine-grained, and Animals With Attributes (AWA) \cite{LNH13} is a standard attribute dataset for zero-shot classification. 
CUB contains 11,788 images of 200 bird species, Dogs contains 19,501 images of 113 dog breeds and AWA contains 30,475 images of 50 different animals. 
We use a truly zero-shot setting where the train, val, and test sets belong to mutually exclusive classes. We employ train and val, \ie disjoint subsets of training set, for cross-validation. We report average per-class top-1 accuracy on the test set. For CUB, we use the same zero-shot split as \cite{APHS13} with 150 classes for the train+val set and 50 disjoint classes for the test set. AWA has a predefined split for 40 train+val and 10 test classes. For Dogs, we use approximately the same ratio of classes for train+val/test as CUB, \ie 85 classes for train+val and 28 classes for test. This is the first attempt to perform zero-shot learning on the Dogs dataset.

\myparagraph{Input Embeddings.} 
We use Fisher Vectors (FV) and Deep CNN Features (CNN). 
FV~\cite{PD07} aggregates per image statistics computed from local image patches into a fixed-length local image descriptor. 
We extract 128-dim SIFT from regular grids at multiple scales, reduce them to 64-dim using PCA, build a visual vocabulary with 256 Gaussians~\cite{VF08} and finally reduce the FVs to 4,096.
As an alternative, we extract features from a deep convolutional network.
Features that are typically obtained from the activations of the fully connected layers have been shown to induce semantic similarities.
We resize each image to 224$\times$224 and feed into the network which was pre-trained following the model architecture of either AlexNet~\cite{KSH12} or GoogLeNet~\cite{szegedy2014going,ioffe2015batch}.
For AlexNet (denoted as CNN) we use the 4,096-dim top-layer hidden unit activations (“fc7”) as features, and for GoogLeNet (denoted as GOOG) we use the 1,024-dim top-layer pooling units.
For both networks, we used the publicly-available BVLC implementations~\cite{jia2014caffe}.
We do not perform any task-specific pre-processing, such as cropping foreground objects or detecting parts.

\myparagraph{Output Embeddings.} 
AWA classes have 85 binary and continuous attributes. CUB classes have 312 continuous attributes and the continuous values are thresholded around the mean to obtain binary attributes. 
The Dogs dataset does not have human-annotated attributes available.

We train Word2Vec ($\p^{\cal W}$) and GloVe ($\p^{\cal G}$) on the English-language \texttt{Wikipedia} from 13.02.2014. 
We first pre-process it by replacing the class-names, \ie \emph{black-footed albatross}, with alternative unique names, \ie scientific name, \emph{phoebastrianigripes}.
We cross-validate the skip-window size and embedding dimensions.
For our proposed weakly-supervised Word2Vec ($\p^{{\cal W}_{ws}}$), we use the same embedding dimensions as the plain Word2Vec ($\p^{\cal W}$).
For BoW, we download the \texttt{Wikipedia} articles that correspond to each class and build the vocabulary by omitting least- and most-frequently occurring words.
We cross-validate the vocabulary size.
$\p^{\cal B}$ is a histogram of the vocabulary words as they appear in the respective document.

For hierarchical embeddings ($\p^{\cal H}$), we use the WordNet hierarchy spanning our classes and their ancestors up to the root of the tree.
We employ the widely used NLTK library\footnote{\texttt{http://www.nltk.org/}} for building the hierarchy and measuring the similarity between nodes.
Therefore, each $\p^{\cal H}$ vector is populated with similarity measures of the class to all other classes.

\myparagraph{Combination of output embeddings.}
We explore combinations of five types of output embeddings: supervised attributes $\p^{\cal A}$, unsupervised Word2Vec $\p^{\cal W}$, GloVe $\p^{\cal G}$, BoW $\p^{\cal B}$ and WordNet-derived similarity embeddings $\p^{\cal H}$.
We either concatenate (\emph{cnc}) or combine (\emph{cmb}) different embeddings. In \emph{cnc}, for instance in AWA, 85-dim $\p^{\cal A}$ and 400-dim $\p^{\cal W}$ would be merged to 485-dim output embeddings. In this case, if we use 1,024-dim GOOG as input embeddings, we learn a single 1,024$\times$485-dim $W$. In \emph{cmb}, we first learn 1,024$\times$85-dim $W_{\cal A}$ and 1,024$\times$400-dim $W_{\cal W}$ and then cross-validate the $\alpha$ coefficients to determine the amount each embedding contributes to the final score.  

\subsection{Experimental Results}
\label{subsec:results}

In this section, we evaluate several output embeddings on the CUB, AWA and Dogs datasets.

\begin{table}[t]
 \begin{center}
   \small
  \begin{tabular}{|c|r|c c |c c |c c|}
	\hline
	 & & \multicolumn{2}{c|}{\textbf{AWA}} & \multicolumn{2}{c|}{\textbf{CUB}} \\ 
	\hline
	 &  & $\p^{{0,1}}$ & $\p^{\cal A}$ & $\p^{{0,1}}$ & $\p^{\cal A}$ \\ 
	\hline
	\multirow{3}{*}{Ours} & FV (4K) & 36.6 & 42.3 & 15.2 & 19.0 \\ 
	& CNN (4K) & 45.9 & 61.9 & 30.0 & 40.3 \\ 
	& GOOG (1K) & 52.0 & \textbf{66.7} & 37.8 & \textbf{50.1} \\
	\hline
	SoA & ALE~\cite{APHS15} (64K)& 44.6 & 48.5 & 22.3 &  26.9 \\
	\hline
  \end{tabular}
 \end{center}
\caption{Discrete ($\p^{0,1}$) and continuous ($\p^{\cal A}$) attributes with \SJE vs SoA. For AWA (CUB) \cite{APHS15} achieves 49.4\% (27.3\%) by combining $\p^{\cal A}$ and binary hierarchies. 
}
\cutcaptiondown
\label{tab:att}
\end{table}

\myparagraph{Discrete vs Continuous Attributes.} 
Attribute representations are defined as a vector per class, or a column of the (class $\times$ attribute) matrix. These vectors (85-dim for AWA, 312-dim for CUB) can either model the presence/absence ($\p^{0,1}$) or the confidence level ($\p^{\cal A}$) of each attribute.
We show that continuous attributes indeed encode more semantics than binary attributes by observing a substantial improvement with $\p^{\cal A}$ over $\p^{0,1}$ with deep features (Tab.~\ref{tab:att}).
Overall, CNN outperforms FV, 
while GOOG gives the best performing results; therefore in the following, we comment only on our results obtained using GOOG.

On CUB, \ie a fine-grained dataset, $\p^{0,1}$ obtains 37.8\% accuracy, which is significantly above the SoA (26.9\%~\cite{APHS15}). 
Moreover, $\p^{\cal A}$ achieves an impressive 50.1\% accuracy; outperforming the SoA by a large margin.
We observe the same trend for AWA, which is a benchmark dataset for zero-shot learning. 
On AWA, $\p^{0,1}$ obtains 52.0\% accuracy and $\p^{\cal A}$ improves the accuracy substantially to 66.7\%, significantly outperforming the SoA (48.5\%~\cite{APHS15}).
To summarize, we have shown that $\p^{\cal A}$ improves the performance of $\p^{0,1}$ using deep features, which indicates that with $\p^{\cal A}$, the \SJE method learns a matrix $W$ that better approximates the compatibility of images and side information than $\p^{0,1}$. 

\begin{table}[t]
 \begin{center}
   \small
  \begin{tabular}{|c|c|c|c|c|c|}
	\hline
	supervision & source & $\p$ & \textbf{AWA} & \textbf{CUB} & \textbf{Dogs} \\
	\hline
	
	\multirow{4}{*}{unsupervised} & text & $\p^{\cal W}$ & 51.2 & \textbf{28.4} & 19.6\\
	 & text & $\p^{\cal G}$ & \textbf{58.8} & 24.2 & 17.8\\
	 & text & $\p^{\cal B}$ & 44.9 & 22.1 & \textbf{33.0} \\
	 & WordNet & $\p^{\cal H}$ & 51.2 & 20.6 & 24.3\\

	\hline
	
	\multirow{2}{*}{supervised} & human & $\p^{0,1}$ & 52.0 & 37.8 & -\\
	& human & $\p^{\cal A}$ & \textbf{66.7} & \textbf{50.1} & -\\
	\hline
  \end{tabular}
 \end{center}
\caption{Summary of zero-shot learning results with \SJE \wrt supervised and unsupervised output embeddings (Input embeddings: 1K-GOOG).}
\cutcaptiondown
\label{tab:summary}
\end{table}

\myparagraph{Learned Embeddings from Text.} 
As the visual similarity between objects in different classes increases, \eg in fine-grained datasets, the cost of collecting attributes also increases. Therefore, we aim to extract class similarities automatically from unlabeled online textual resources. 
We evaluate three methods, Word2Vec ($\p^{\cal W}$), GloVe ($\p^{\cal G}$) and the historically most commonly-used method BoW ($\p^{\cal B}$). We build $\p^{\cal W}$ and $\p^{\cal G}$ on the entire English Wikipedia dump. Note that the plain Word2Vec~\cite{MSCCD13} was used in~\cite{APHS15}; however, rather than using Word2Vec in an averaging mechanism, we pre-process the Wikipedia as described in Sec~\ref{subsec:text} so that our class names are directly present in the Word2Vec vocabulary. This leads to a significant accuracy improvement. For $\p^{\cal B}$ we use a subset of Wikipedia populated only with articles that correspond to our classes. 
On CUB (Tab.~\ref{tab:summary}), the best accuracy is observed with $\p^{\cal W}$ (28.4\%) improving the supervised SoA (26.9\%~\cite{APHS15}, Tab.~\ref{tab:att}).
This is promising and impressive since $\p^{\cal W}$ does not use any human supervision.
On AWA (Tab.~\ref{tab:summary}), the best accuracy is observed with $\p^{\cal G}$ (58.8\%) followed by $\p^{\cal W}$ (51.2\%), improving the supervised SoA (48.5\%~\cite{APHS15}) significantly.
On Dogs (Tab.~\ref{tab:summary}), the best accuracy is obtained with $\p^{\cal B}$ (33.0\%). On the other hand, using $\p^{\cal W}$ (19.6\%) and $\p^{\cal G}$ (17.8\%) leads to significantly lower accuracies. 
Unlike birds, different dog breeds belong to the same species and thus they share a common scientific name. As a result, our method of cleanly pre-processing Wikipedia by replacing the occurrences of bird names with a unique scientific name was not  possible for Dogs. This may lead to vectors obtained from Wikipedia for dogs that are vulnerable to variation in nomenclature.
In summary, our results indicate no winner among $\p^{\cal W}$, $\p^{\cal G}$ and $\p^{\cal B}$. These embeddings may be task specific and complement each other. We investigate the complementarity of embeddings in the following sections.

\begin{table}[t]
 \begin{center}
  \resizebox{\linewidth}{!}{
  \begin{tabular}{|r|c c c |c c c | c|}
	\hline
	& \multicolumn{3}{c|}{$\p^{\cal G}$} & \multicolumn{3}{c|}{$\p^{\cal W}$} & $\p^{\cal W}$ (W) + \\

	& B & W & B+W & B & W & B+W & $\p^{{\cal W}_{ws}}$ (B) \\
	\hline
	FV & 10.5 & 13.3 & 13.2  & 16.0 & 16.0 & 16.5 & 17.1 \\
	CNN & 13.4 & 20.6 & 20.6 &  20.0 & 24.1 &  21.4 & 25.1 \\
	GOOG & 13.7 & 24.2 & \textbf{26.1} & 22.5 & \textbf{28.4} & 27.5 & \textbf{29.7}\\
	\hline
  \end{tabular}
}
 \end{center}
\caption{Comparison of Word2Vec ($\p^{\cal W}$) and GloVe ($\p^{\cal G}$) learned from a bird specific corpus (B), Wikipedia (W) and their combination (B + W), evaluated on CUB (Input embeddings: 4K-FV, 4K-CNN and 1K-GOOG).}
\cutcaptiondown
\label{tab:text_corpora}
\end{table}

\myparagraph{Effect of Text Corpus.} 
For $\p^{\cal W}$ and $\p^{\cal G}$, we analyze the effects of three text corpora (B, W, B+W) with varying size and specificity.
We build our specialized bird corpus (B) by collecting bird-related information from various online resources, \ie \texttt{audubon.org}, \texttt{birdweb.org}, \texttt{allaboutbirds.org} and BNA\footnote{\texttt{http://bna.birds.cornell.edu/bna/}}. In combination, this corresponds to 50MB of bird-related text.
We use the English-language \texttt{Wikipedia} from 13.02.2014 as our large and general corpus (W) which is 40GB of text.
Finally, we combine B and W to build a large-scale text corpus enriched with bird specific text (B+W).
On W and B+W, a small window size (10 for $\p^{\cal W}$ and 20 for $\p^{\cal G}$); on B, a large window size (35 for $\p^{\cal W}$ and 50 for $\p^{\cal G}$) is required. We choose parameters after a grid search. 
Increased specificity of the text corpus implies semantic consistency throughout the text. Therefore, large context windows capture semantics well in our bird specific (B) corpus. On the other hand, W is organized alphabetically \wrt the document title; hence, a large sampling window can include content from another article that is adjacent to the target word alphabetically. Here, small windows capture semantics better by looking at the text locally. We report our results in Tab.~\ref{tab:text_corpora}.

Using $\p^{\cal G}$, B+W (26.1\%) gives the highest accuracy, followed by W (24.2\%).
One possible reason is that when the semantic similarity is modeled with cooccurrence statistics, output embeddings become more informative with the increasing corpus size, since the probability of cooccurrence of similar concepts increases. 

Using $\p^{\cal W}$, the accuracy obtained with B (22.5\%) is already higher than the  $\p^{0,1}$-based SoA (22.3\%), illustrating the benefit of using fine-grained text for fine-grained tasks. Another advantage of using B is that, since it is short, building $\p^{\cal W}$ is efficient. 
Moreover, building $\p^{\cal W}$ with B does not require any annotation effort. Building $\p^{\cal W}$ using W (28.4\%) gives the highest accuracy, followed by W + B (27.5\%) which improves the supervised SoA (26.9\%). 
We speculate that since Word2Vec is a variant of the Feedforward Neural Network Language Model (FNNLM)~\cite{BDVJ03}, a deep architecture, it may learn more from negative data than positives.
This was also observed for CNN features learned with a large number of unlabeled surrogate classes \cite{DSRB14}. 

Additionally, we propose a weakly-supervised alternative to Word2Vec framework ($\p^{{\cal W}_{ws}}$, Sec.~\ref{subsec:text}). The weak-supervision comes from using the specialized B corpus to fine-tune the weights of the network and model the bird-related information.
With $\p^{{\cal W}_{ws}}$ alone, we obtain 21.0\% accuracy.
However, when it is combined with $\p^{\cal W}$ (28.4\%), the accuracy improves to 29.7\%. Compared to the results in Tab.~\ref{tab:text_corpora}, 29.7\% is the highest accuracy obtained using unsupervised embeddings. We regard these results as a very encouraging evidence that Word2Vec representations can indeed be made more discriminative for fine-grained zero-shot learning by integrating a fine-grained text corpus directly to the output embedding learning problem.

\begin{figure}[t]
   \centering
   \includegraphics[width=\linewidth,trim=40 400 40 80]{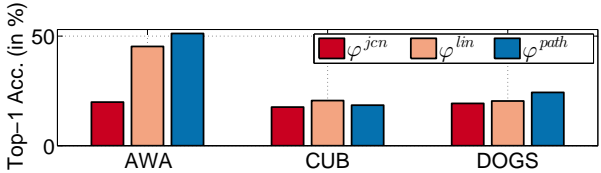}
\caption{Comparison of WordNet similarity measures: $\p^{jcn}$, $\p^{lin}$ and $\p^{path}$. We use $\p^{\cal H}$ as a general name for hierarchical output embedding. (Input embedding: 1K-GOOG).} 
\cutcaptiondown
\label{fig:hie}
\end{figure}

\myparagraph{Hierarchical Embeddings.}  
The hierarchical organization of concepts typically embodies a fair amount of hidden information about language, such as synonymy, semantic relations, \etc. Therefore, semantic relatedness defined by hierarchical distance between classes can form numerical vectors to be used as output embeddings for zero-shot learning. 
We build ontological relationships between our classes using the WordNet~\cite{WordNet} taxonomy. Due to its large size, WordNet encapsulates all of our AWA and Dog classes.
For CUB, the high level bird species, \ie albatross, appear as synsets in WordNet, but the specific bird names, \ie black-footed albatross, are not always present.
Therefore we take the hierarchy up to high level bird species as-is and we assume the specific bird classes are all at the bottom of the hierarchy located with the same distance to their immediate ancestors. The WordNet hierarchy contains 319 nodes for CUB (200 classes), 104 nodes for AWA (50 classes) and 163 nodes for Dogs (113 classes).
We measure the distance between classes using the similarity measures from Sec~\ref{subsec:attributes}. 

While as shown in Fig.~\ref{fig:hie} different hierarchical similarity measures have very different behaviors on each dataset. 
The best performing $\p^{\cal H}$ obtains 51.2\% (Tab.~\ref{tab:summary}) accuracy on AWA which reaches our $\p^{0,1}$ (52.0\%) and improves $\p^{\cal B}$ (44.9\%) significantly. On CUB, $\p^{\cal H}$ obtains 20.6\% (Tab.~\ref{tab:summary}) which remain below our $\p^{0,1}$ (37.8\%) and approaches $\p^{\cal B}$ (22.1\%). On the other hand, on Dogs $\p^{\cal H}$ obtains 24.3\% (Tab.~\ref{tab:summary}) which is significantly higher than the unsupervised text embeddings $\p^{\cal W}$ (19.6\%) and $\p^{\cal G}$ (17.8\%).

\newcommand{\vv}{\checkmark}

\begin{table}[t]
 \begin{center}
\resizebox{\linewidth}{!}{
  \begin{tabular}{|m{0.2cm} m{0.2cm} m{0.2cm} m{0.2cm} m{0.2cm} | c c | c c | c c |}
	\hline
	& & & & & \multicolumn{2}{c|}{\textbf{AWA}} & \multicolumn{2}{c|}{\textbf{CUB}} & \multicolumn{2}{c|}{\textbf{Dogs}} \\
	\hline
	$\p^{\cal A}$ & $\p^{\cal W}$ & $\p^{\cal G}$ & $\p^{\cal B}$ & $\p^{\cal H}$ & \emph{cnc} & \emph{cmb} & \emph{cnc} & \emph{cmb} & \emph{cnc} & \emph{cmb}  \\
	\hline
	  & $\vv$ &       &       & $\vv$ & 53.9 & 55.5 & 28.2 & 29.4 & 23.5 &  26.6 \\ 
	  &       & $\vv$ &       & $\vv$ & \textbf{60.1} & 59.5 & 28.5 & \textbf{29.9} & 23.5 & 26.7 \\ 
	  &       &       &  $\vv$& $\vv$ & 49.4 & 49.2 & 26.4 & 27.7 & \textbf{35.1}  &  28.2 \\ 
	\hline

	$\vv$ & $\vv$ &       &       &  $\vv$& 71.3 & 73.5 & 45.1 & 51.0 & - & -  \\
	$\vv$ &       & $\vv$ &       &  $\vv$& 73.3 & \textbf{73.9} & 42.2 & \textbf{51.7} & - & - \\
	$\vv$ &       &       & $\vv$ &  $\vv$& 69.4 & 71.1 & 40.9 & 51.5 & - & - \\
	\hline
  \end{tabular}
}
\end{center}
\caption{Attribute ensemble results for all datasets. $\p^{\cal H}$: lin for CUB, path for AWA and Dogs. Top part shows combination results of unsupervised embeddings and bottom part integrates supervised embeddings to the rest (Input embeddings: 1K-GOOG).}
\cutcaptiondown
\label{tab:ensemble_all}
\end{table}
\begin{figure*}[t]
\begin{center}
\includegraphics[width=\linewidth, , trim=30 50 30 30]{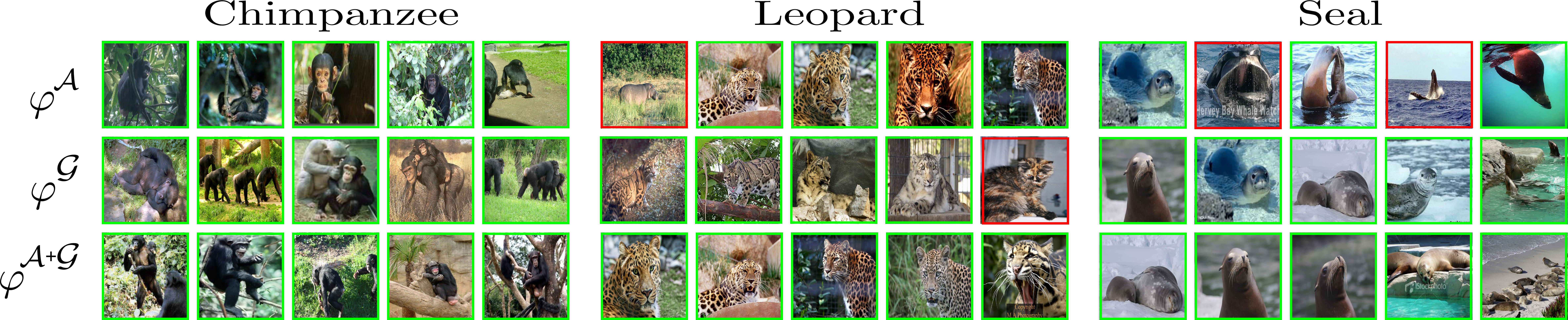}
\end{center}
\caption{Highest ranked 5 images for \emph{chimpanzee}, \emph{leopard} and \emph{seal} (AWA) using $\p^{\cal A}$, $\p^{\cal G}$ and $\p^{\cal G+A}$. For \emph{chimpanzee}, $\p^{\cal A}$ ranks chimpanzees on trees at the top, whereas $\p^{\cal G}$ models the social nature of the animal ranking a group of chimpanzees highest, $\p^{\cal G+A}$ synthesizes both aspects. For \emph{leopard} $\p^{\cal A}$ puts an emphasis on the head, $\p^{\cal G}$ seems to place the animal in the wild. In case of \emph{seal}, $\p^{\cal A}$ retrieves images related to \emph{water}, whereas $\p^{\cal G}$ adds more context by placing seals in the icy natural environment and $\p^{\cal G+A}$ combines both.}
\cutcaptiondown

\label{fig:toprankedimg}
\end{figure*}
%

\myparagraph{Combining Output Embeddings.}
In this section, we combine output embeddings obtained through human annotation ($\p^{\cal A}$), from text ($\p^{\cal W,G,B}$) and from hierarchies ($\p^{\cal H}$).\footnote{We empirically found that the hierarchical embeddings $\p^{\cal H}$ consistently improved performance when combined or concatenated with other embeddings. Therefore, we report results using $\p^{\cal H}$ by default.}
As a reference, Tab.~\ref{tab:summary} summarizes the results obtained using one output embedding at a time. Our intuition is that because the different embeddings attempt to encapsulate different information, accuracy should improve when multiple embeddings are combined. 
We can observe this complementarity either by simple concatenation (\emph{cnc}) or systematically combining (\emph{cmb}) output embeddings (Sec.\ref{subsec:comb_emb}) also known as early/late fusion~\cite{APHS15}. 
For \emph{cnc}, we perform full SJE training and cross-validation on the concatenated output embeddings.
For \emph{cmb}, we learn joint embeddings $W_{k}$ for each output separately (which is trivially parallelized), and find ensemble weights $\alpha_{k}$ via cross-validation.
In contrast to the \emph{cnc} method, no additional joint training is used, although it can improve performance in practice. 
We observe (Tab.~\ref{tab:ensemble_all}) in almost all cases \emph{cmb} outperforms \emph{cnc}.

We analyze the combination of unsupervised embeddings ($\p^{\cal W,G,B,H}$). On AWA, $\p^{\cal G}$ (58.8\%, Tab.~\ref{tab:summary}) combined with $\p^{\cal H}$ (51.2\%, Tab.~\ref{tab:summary}), 
we achieve 60.1\% (Tab.~\ref{tab:ensemble_all}) which improves the SoA (48.5\%, Tab.~\ref{tab:att}) by a large margin.
On CUB, combining $\p^{\cal G}$ (24.2\%, Tab.~\ref{tab:summary}) with $\p^{\cal H}$ (20.6\%, Tab.~\ref{tab:summary}), 
we get 29.9\% (Tab.~\ref{tab:ensemble_all}) 
and improve the supervised-SoA  (26.9\%, Tab.~\ref{tab:att}).
Supporting our initial claim, unsupervised output embeddings obtained from different sources, \ie text vs hierarchy, seem to be complementary to each other.
In some cases, \textit{cmb} performs worse than \textit{cnc}; \eg 28.2\% versus 35.1\% when using $\p^{\cal B}$ with $\p^{\cal H}$ on Dogs.
In most other cases \textit{cmb} performs equivalent or better.
Combining supervised ($\p^{\cal A}$) and unsupervised embeddings ($\p^{\cal W,G,B,H}$) shows a similar trend. On AWA, combining $\p^{\cal A}$ (66.7\%, Tab.~\ref{tab:summary}) with $\p^{\cal G}$ and $\p^{\cal H}$ leads to 73.9\% (Tab.~\ref{tab:ensemble_all}) which significantly exceeds the SoA (48.5\%, Tab.~\ref{tab:att}). On CUB, combining $\p^{\cal A}$ with $\p^{\cal G}$ and  $\p^{\cal H}$ leads to 51.7\% (Tab.~\ref{tab:ensemble_all}),
 improving both the results we obtained with $\p^{\cal A}$ (50.1\%, Tab.~\ref{tab:summary}) and the supervised-SoA (26.9\%, Tab.~\ref{tab:att}). We have shown with these experiments that output embeddings obtained through human annotation can also be complemented with unsupervised output embeddings using the \SJE framework.

\begin{table}[t]
 \begin{center}
   \small
  \begin{tabular}{c|c|c|c|c}
	\hline
	supervision & method &\textbf{AWA} & \textbf{CUB} & \textbf{Dogs} \\
	\hline
	unsupervised & \SJE (best from Tab.~\ref{tab:ensemble_all}) & 60.1 & 29.9 & \textbf{35.1}\\
	\hline
	\multirow{2}{*}{supervised} & \SJE (best from Tab.~\ref{tab:ensemble_all}) & \textbf{73.9} & \textbf{51.7} & -- \\
	 & AHLE~\cite{APHS15} & 49.4 & 27.3 & -- \\
	\hline
  \end{tabular}
 \end{center}
\caption{Summary of best zero-shot learning results with \SJE with or without supervision along with SoA.} \vspace{-3mm} 
\cutcaptiondown
\label{tab:summary_all}
\end{table}

\myparagraph{Qualitative Results.} Fig.~\ref{fig:toprankedimg} shows top-5 highest ranked images for classes \emph{chimpanzee}, \emph{leopard} and \emph{seal} that are selected from 10 test classes of AWA. We use GOOG as input embeddings and as output embeddings we use supervised $\p^{\cal A}$, the best performing unsupervised embedding on AWA ($\p^{\cal G}$), and the combination of the two ($\p^{\cal G+A}$). For the class \emph{chimpanzee}, $\p^{\cal A}$ emphasizes that chimpanzees live on trees, which is among the list of attributes. On the other hand, $\p^{\cal G}$ models the social nature of the animal, ranking a group of chimpanzees interacting with each other at the highest. Indeed this information can easily be retrieved from Wikipedia. $\p^{\cal G+A}$ synthesizes both aspects. Similarly, for \emph{leopard} $\p^{\cal A}$ puts an emphasis on the head where we can observe several of the attributes, \ie color, spotted, whereas $\p^{\cal G}$ seems to place the animal in the wild. $\p^{\cal G+A}$ combines both aspects. In case of class \emph{seal}, $\p^{\cal A}$ retrieves images related to \emph{water} and ranks whales and seals highest, whereas $\p^{\cal G}$ adds more context by placing seals in the icy natural environment and within groups. Finally, $\p^{\cal G+A}$ ranks seal-shaped animals on ice, close to water and within groups the highest. We find these qualitative results interesting as they depict how (1) unsupervised embeddings capture nameable semantics about objects and (2) different output embeddings are semantically complementary for zero-shot learning.  

%
%
\section{Conclusion}
\label{sec:conclusion}
We evaluated the \SJElong (SJE) framework on supervised attributes and unsupervised output embeddings obtained from hierarchies and unlabeled text corpora. We proposed a novel weakly-supervised label embedding technique. 
By combining multiple output embeddings (\emph {cmb}), we established a new SoA on AWA (73.9\%, Tab.~\ref{tab:summary_all}) and CUB (51.7\%, Tab.~\ref{tab:summary_all}).
Moreover, we showed that unsupervised zero-shot learning with \SJE improves the SoA, to 60.1\% on AWA and 29.9\% on CUB, and obtains 35.1\% on Dogs (Tab.~\ref{tab:summary_all}). 

We emphasize the following take-home points: (1) Unsupervised label embeddings learned from text corpora yield compelling zero-shot results, outperforming previous supervised SoA on AWA and CUB (Tab.~\ref{tab:att} and~\ref{tab:summary}).
(2) Integrating specialized text corpora helps due to incorporating more fine-grained information to output embeddings (Tab.~\ref{tab:text_corpora}).
(3) Combining unsupervised output embeddings improve the zero-shot performance, suggesting that they provide complementary information (Tab.~\ref{tab:ensemble_all}).
(4) There is still a large gap between the performance of unsupervised output embeddings and human-annotated attributes on AWA and CUB, suggesting that better methods are needed for learning discriminative output embeddings from text. (5) Finally, supporting~\cite{APHS15, RSSS12}, encoding continuous nature of attributes significantly improve upon binary attributes for zero-shot classification (Tab.~\ref{tab:att}).

As future work, we plan to investigate other methods to combine multiple output embeddings and to improve the discriminative power of unsupervised and weakly-supervised label embeddings for fine-grained classification.

\subsection*{Acknowledgments}
\vspace*{-0.05in}
This work was supported in part by ONR N00014-13-1-0762, NSF CMMI-1266184, Google Faculty Research Award, and NSF Graduate Fellowship.

{\small
\bibliographystyle{ieee}
\bibliography{myrefs}
}

\end{document}